\documentclass[10pt,twocolumn,letterpaper]{article}

\usepackage{cvpr}
\usepackage{times}
\usepackage{epsfig}
\usepackage{graphicx}
\usepackage{amsmath}
\usepackage{amssymb}

\usepackage{bm}
\usepackage{amsmath}
\usepackage{algorithm}
\usepackage{algorithmic}
\usepackage{amssymb}
\usepackage{booktabs}
\usepackage{subfigure}
\usepackage{textcomp}
\usepackage{multirow}
\usepackage{wrapfig}
\usepackage{caption}
\usepackage{comment}
\usepackage{amsthm}
\newtheorem{theorem}{Theorem}

\usepackage{comment}
\usepackage{mathrsfs}

\usepackage[breaklinks=true,bookmarks=false]{hyperref}

\cvprfinalcopy 

\def\cvprPaperID{4813} 

\DeclareMathOperator*{\argmax}{\arg\max}
\DeclareMathOperator*{\argmin}{\arg\min}
\DeclareMathOperator*{\Exp}{\mathbb{E}}

\newcommand{\tabincell}[2]{\begin{tabular}{@{}#1@{}}#2\end{tabular}}  

\begin{document}

\newcommand{\fracpartial}[2]{\frac{\partial #1}{\partial  #2}}
\newcommand{\norm}[1]{\left\lVert#1\right\rVert}
\newcommand{\innerproduct}[2]{\left\langle#1, #2\right\rangle}
\newcommand{\fan}[1]{\Vert #1 \Vert}
\newcommand{\qileft}{[\kern-0.15em[}
\newcommand{\qiLeft}{\left[\kern-0.4em\left[}
\newcommand{\qiright}{]\kern-0.15em]}
\newcommand{\qiRight}{\right]\kern-0.4em\right]}
\newcommand{\sign}{{\mbox{sign}}}
\newcommand{\diag}{{\mbox{diag}}}
\newcommand{\armin}{{\mbox{argmin}}}
\newcommand{\rank}{{\mbox{rank}}}
\renewcommand{\vec}{{\mbox{vec}}}
\newcommand{\st}{{\mbox{s.t.}}}
\newcommand{\<}{\left\langle}
\renewcommand{\>}{\right\rangle}
\newcommand{\lbar}{\left\|}
\newcommand{\rbar}{\right\|}
\renewcommand{\algorithmicrequire}{\textbf{Input:}} 
\renewcommand{\algorithmicensure}{\textbf{Output:}} 
\renewcommand{\Roman}[1]{\uppercase\expandafter{\romannumeral#1}}
\newcommand{\red}[1]{{\color{red}{#1}}}
\newcommand{\blue}[1]{{\color{blue}{#1}}}

\renewcommand{\a}{{\bm{a}}}
\renewcommand{\b}{{\bm{b}}}
\renewcommand{\d}{{\bm{d}}}
\newcommand{\e}{{\bm{e}}}
\newcommand{\f}{{\bm{f}}}
\newcommand{\g}{{\bm{g}}}
\renewcommand{\o}{{\bm{o}}}
\newcommand{\p}{{\bm{p}}}
\newcommand{\q}{{\bm{q}}}
\renewcommand{\r}{{\bm{r}}}
\newcommand{\s}{{\bm{s}}}
\renewcommand{\t}{{\bm{t}}}
\renewcommand{\u}{{\bm{u}}}
\renewcommand{\v}{{\bm{v}}}
\newcommand{\w}{{\bm{w}}}
\newcommand{\x}{{\bm{x}}}
\newcommand{\y}{{\bm{y}}}
\newcommand{\z}{{\bm{z}}}
\newcommand{\balpha}{{\bm{\alpha}}}
\newcommand{\bbeta}{{\bm{\beta}}}
\newcommand{\bmu}{{\bm{\mu}}}
\newcommand{\bsigma}{{\bm{\sigma}}}
\newcommand{\blambda}{{\bm{\lambda}}}
\newcommand{\bgamma}{{\bm{\gamma}}}
\newcommand{\bxi}{{\bm{\xi}}}
\newcommand{\bphi}{{\bm{\phi}}}

\newcommand{\ba}{{\bm{A}}}
\newcommand{\bb}{{\bm{B}}}
\newcommand{\bc}{{\bm{C}}}
\newcommand{\bd}{{\bm{D}}}
\newcommand{\be}{{\bm{E}}}
\newcommand{\bg}{{\bm{G}}}
\newcommand{\bi}{{\bm{I}}}
\newcommand{\bj}{{\bm{J}}}
\newcommand{\bl}{{\bm{L}}}
\newcommand{\bo}{{\bm{O}}}
\newcommand{\bp}{{\bm{P}}}
\newcommand{\bq}{{\bm{Q}}}
\newcommand{\bs}{{\bm{S}}}
\newcommand{\bu}{{\bm{U}}}
\newcommand{\bv}{{\bm{V}}}
\newcommand{\bw}{{\bm{W}}}
\newcommand{\bx}{{\bm{X}}}
\newcommand{\by}{{\bm{Y}}}
\newcommand{\bz}{{\bm{Z}}}
\newcommand{\bTheta}{{\bm{\Theta}}}
\newcommand{\bSigma}{{\bm{\Sigma}}}

\newcommand{\A}{{\mathcal{A}}}
\newcommand{\B}{\mathcal{B}}
\newcommand{\C}{\mathcal{C}}
\newcommand{\D}{\mathcal{D}}
\newcommand{\F}{\mathcal{F}}
\renewcommand{\H}{\mathcal{H}}
\newcommand{\I}{\mathcal{I}}
\renewcommand{\L}{\mathcal{L}}
\newcommand{\N}{\mathcal{N}}
\renewcommand{\P}{\mathcal{P}}
\newcommand{\X}{\mathcal{X}}
\newcommand{\Y}{\mathcal{Y}}
\newcommand{\W}{\mathcal{W}}

\title{GreedyNAS: Towards Fast One-Shot NAS with Greedy Supernet}
\author{Shan You$^{1,2}$\thanks{Equal contributions.}, Tao Huang$^{1,3*}$, Mingmin Yang$^{1*}$, Fei Wang$^{1}$, Chen Qian$^{1}$, Changshui Zhang$^{2}$\\
$^1$SenseTime   \quad $^2$Department of Automation, Tsinghua University\\
$^3$Dian Group, School of CST, Huazhong University of Science and Technology\\
{\tt\small \{youshan,huangtao,yangmingmin,wangfei,qianchen\}@sensetime.com zcs@mail.tsinghua.edu.cn}
}

\maketitle

\begin{abstract}
	Training a supernet matters for one-shot neural architecture search (NAS) methods since it serves as a basic performance estimator for different architectures (paths). Current methods mainly hold the assumption that a supernet should give a reasonable ranking over all paths. They thus treat all paths equally, and spare much effort to train paths. However, it is harsh for a single supernet to evaluate accurately on such a huge-scale search space (\eg, $7^{21}$). In this paper, instead of covering all paths, we ease the burden of supernet by encouraging it to focus more on evaluation of those potentially-good ones, which are identified using a surrogate portion of validation data. Concretely, during training, we propose a multi-path sampling strategy with rejection, and greedily filter the weak paths. The training efficiency is thus boosted since the training space has been greedily shrunk from all paths to those potentially-good ones. Moreover, we further adopt an exploration and exploitation policy by introducing an empirical candidate path pool. Our proposed method GreedyNAS is easy-to-follow, and experimental results on ImageNet dataset indicate that it can achieve better Top-1 accuracy under same search space and FLOPs or latency level, but 
	with only $\sim$60\% of supernet training cost. By searching on a larger space, our GreedyNAS can also obtain new state-of-the-art architectures.

	\vspace{-5mm}

\end{abstract}

\section{Introduction}
By dint of automatic feature engineering, deep neural networks (DNNs) have achieved remarkable success in various computer vision tasks, such as image classification \cite{wang2017residual,wang2018devil,you2017learning,tang2019bringing,rebornfilters,fewdata,wu2019deep}, visual generation \cite{wang2018perceptual,wang2019evolutionary}, image retrieval \cite{yang2018shared,deng2019unsupervised,deng2019two,han2018attribute} and semantic comprehension \cite{liao2019ppdm,liao2019real}. In contrast, neural architecture search (NAS) aims at automatically learning the network architecture to further boost the performance for target tasks \cite{guo2019irlas,liu2018progressive,ZophL17,chen2019detnas,lin2019graph}. 
Nevertheless, previous NAS methods in general suffer from huge computation budget, such as 2000 GPU days of reinforcement learning \cite{ZophL17} and 3150 GPU days of evolution \cite{real2019regularized} with hundreds of GPUs. 

Current One-shot NAS methods boost the search efficiency by modeling NAS as a one-shot training process of an over-parameterized supernet. As a result, various architectures can be derived from the supernet, and share the same weights. For example, DARTS \cite{darts} and its variants 
\cite{fbnet,proxylessnas} parameterize the supernet with an additional categorical distribution for indicating what operations we want to keep. In contrast, recent single path methods adopt a non-parametric architecture modeling, and split the searching into two consecutive stages, \ie, supernet training and architecture sampling. For training supernet, only a single path consisting of a single operation choice is activated and gets optimized by regular gradient-based optimizers. After the supernet is trained well, it is regarded as a performance estimator for all architectures (\ie, paths). Then the optimal architecture can be searched using a hold-out validation dataset via random search \cite{random} or (reinforced) evolutionary  \cite{face++,fairnas} algorithms under specified hardware constraint (\eg, FLOPs and latency). As only one path is activated for training, the memory cost coheres with that of traditional network training, and scales well on large-scale datasets (\eg, ImageNet \cite{imagenet}).

\begin{figure}
	\centering
	\includegraphics[width=\linewidth]{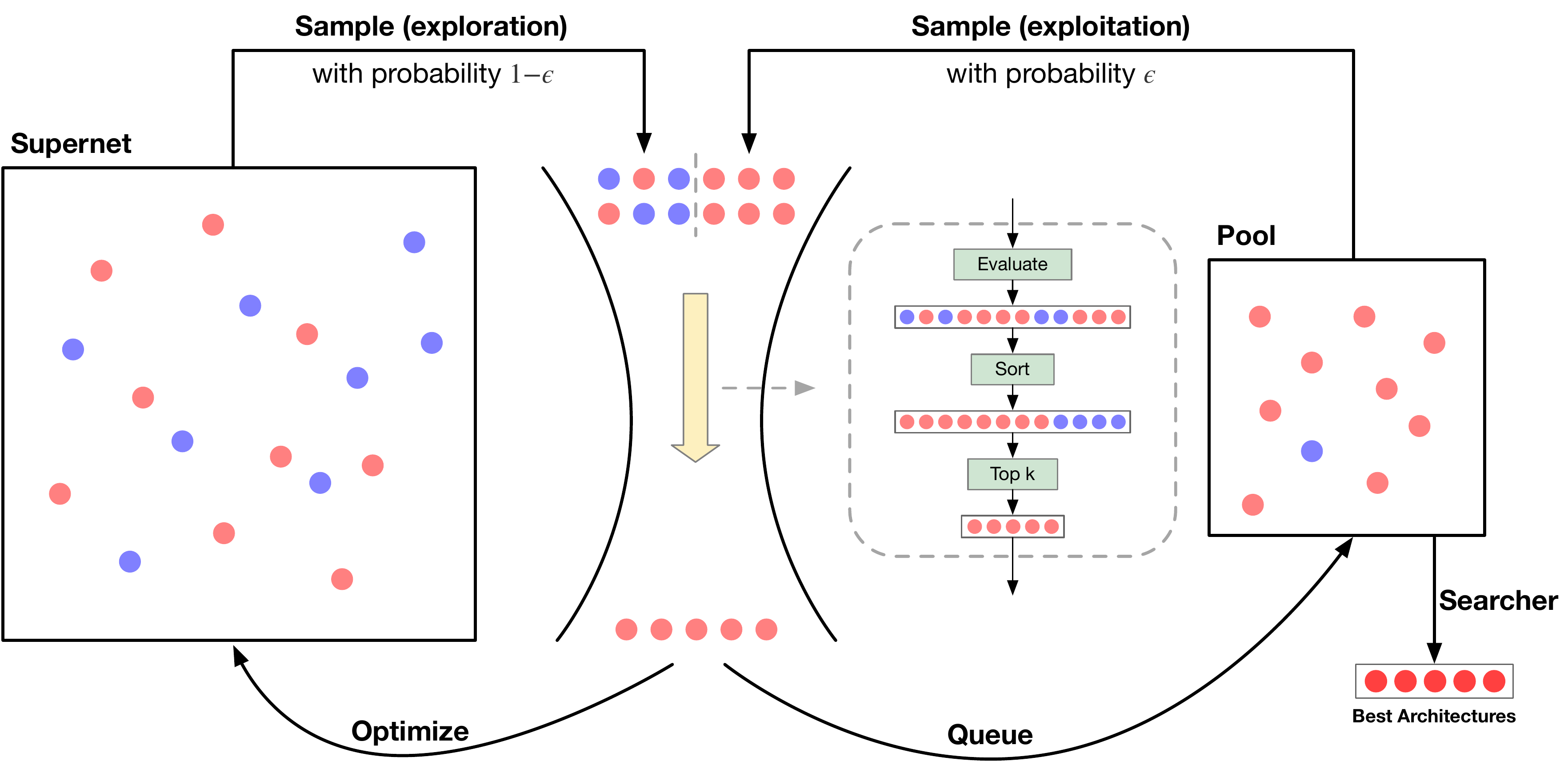}
	\vspace{-5mm}
	\caption{Diagram of supernet training for our proposed GreedyNAS. The supernet greedily shrinks its training space from all paths (\red{red} and \blue{blue} dots) into potentially-good paths (\red{red} dots), and further into candidate pool.}
	\vspace{-4mm}
	\label{fig:shiyi}
\end{figure}

Supernet matters for it serves as a fundamental performance estimator of different architectures (paths). Current methods \cite{random,face++,fairnas,scarletnas} hold the assumption that the supernet should estimate the (relative) performance accurately for \textit{all} paths, and thus all paths are treated equally and trained simultaneously. However, the paths contained in the supernet are of fairly huge scale (\eg, $7^{21}$).  Hence it can be harsh for a single supernet to evaluate and give reasonable ranking on such a quantity of paths at the same time. In fact, the ultimate aim of supernet is only to identify a bunch of optimal paths. But the huge search space implies significant variance and variety of paths; there exist many architectures of inferior quality in terms of accuracy performance.\footnote{For example, in a same supernet, MobileNetV2 \cite{mobilenetv2} can achieve 72.0\% Top-1 accuracy on ImageNet dataset while an extreme case of almost all identity operations only has 24.1\% \cite{scarletnas}.} Since the weights of all paths are highly shared, if a weak path is sampled and gets trained, it would disturb the weights of those potentially-good paths. This disturbance will undermine their eventual performance estimation and affect the searched optimal architecture accordingly. The supernet is thus not supposed to care much on these weak paths and get updated for them. Besides, training on those weak paths actually involves unnecessary update of weights, and slows down the training efficiency more or less. 

In this paper, we ease the training burden by encouraging a greedy supernet. A greedy supernet is capable of shifting its focus on performance estimation of those potentially-good paths instead of all paths. 
Concretely, during the supernet training, we propose a multi-path sampling strategy with rejection to filter the weak paths, so the supernet will greedily train those potentially-good paths. This path filtering can be efficiently implemented via evaluation using a surrogate portion of validation dataset, without harming the computation cost too much. Moreover, we also adopt an exploration and exploitation policy \cite{kocsis2006bandit,mnih2013playing} by introducing a candidate pool, which dynamically tracks those potentially-good paths discovered during training. In this way, the supernet improves its training efficiency by switching its training space from all paths into those potentially-good ones, and further into candidate pool by sampling from it, as shown in Figure \ref{fig:shiyi}.

We implement our proposed method GreedyNAS on the large-scale benchmark ImageNet dataset \cite{imagenet}, and extensive experimental results indicate our superiority in terms of accuracy performance and supernet training efficiency. For example, with the same search space, our method can achieve higher Top-1 accuracy than that of other comparison methods under the same FLOPs or latency level, but reduces approximate 40\% of supernet training cost. By searching on a larger space, we can also obtain new state-of-the-art architectures.

\section{Related Work}
One-shot NAS methods mainly aim to train an over-parameterized network (a.k.a supernet) that comprises all architectures (paths), which share the same weights mutually. Then the optimal architecture can be derived or searched from the supernet. There are mainly two categories of one-shot NAS methods \cite{elsken2019neural}, which differ in how the architectures are modeled and elaborated as follows.

\textbf{Parameterized architectures.} To use the gradient-based optimizers for direct searching, an real-valued categorical distribution (architecture parameter) is usually introduced in the supernet, and can be thus jointly learned with the supernet weights, such as DARTS \cite{darts}, FBNet \cite{fbnet} and MdeNAS \cite{mdenas}. When the supernet training is finished, the optimal architecture can be induced by sampling from the categorical distribution. However, it may suffer from the huge GPU memory consumption. ProxylessNAS \cite{proxylessnas} alleviates this issue by factorizing the searching into multiple binary selection tasks while \textit{Single-Path}-NAS \cite{singlepathnas} uses superkernels to encode all operation choices. Basically, they are difficult to integrate a hard hardware constraint (\eg, FLOPs and latency) during search but resort to relaxed regularization terms \cite{fbnet,proxylessnas}. 

\textbf{Sampled single-path architectures.} By directly searching the discrete search space, the supernet is trained by sampling and optimizing a single path. The sampling can be uniform sampling \cite{face++,random} or multi-path sampling with fairness \cite{fairnas}. After the supernet is trained, it is supposed to act as a performance estimator for different paths. And the optimal path can be searched by various searchers, such as random search and evolutionary algorithms \cite{deb2002fast}. For example, ScarletNAS \cite{scarletnas} employs a multi-objective searcher \cite{lu2018nsga} to consider classification error, FLOPs and model size for better paths. Different to the previous parameterized methods, the hard hardware constraint can be easily integrated in the searchers. Our proposed method GreedyNAS is cast into this category.

\section{Rethinking path training of supernet} 
In Single-path One-shot NAS, we utilize an over-parameterized supernet $\mathcal{N}$ with parameter $\Omega$ to substantialize a search space, which is formulated as a directed acyclic graph (DAG). In the DAG, feature maps act as the nodes, while the operations (or transformations) between feature maps are regarded as edges for connecting sequential nodes. Assume the supernet $\mathcal{N}$ has $L$ layers, and each layer $\mathcal{N}^l$ is allocated with $O$ operation choices $\mathcal{O}=\{o_i\}$, which can be basic convolution, pooling, identity or different types of building blocks, such as MobileNetV2 block \cite{mobilenetv2} and ShuffleNetV2 block \cite{shufflenetv2}. Then each architecture (\ie, path) denoted as $\a$ can be represented by a tuple of size $L$, \ie, $\a = (o^1, o^2, ..., o^L)$ where $o^j \in \mathcal{O},~\forall j = 1,2,...,L$. As a result, the search space $\A$ is discrete, and there will be $O^L$ (\eg, $7^{21}$) architectures in total, namely, $|\A| = O^L$. 

Training supernet matters since it is expected to serve as a fundamental performance estimator. Due to the consideration of memory consumption, single-path NAS methods implement training by sampling a single path $\a$ from $\A$, then the sampled paths are all optimized on the training data $\D_{tr}$. It can be formulated as minimizing an expected loss over the space $\A$, \ie, 
\vspace{-1mm}
\begin{equation}
\Omega^* = \argmin_\Omega  \Exp_{\a\sim p(\A)} \qiLeft \L (\omega_\a;\D_{tr}) \qiRight,
\vspace{-2mm}
\end{equation}
where $\omega_\a$ refers to the parameter of path $\a$, and $p(\A)$ is a discrete sampling distribution over $\A$. 

After the supernet $\N(\Omega^*)$ is trained well, we can evaluate the quality of each path by calculating its (Top-1) accuracy (ACC) on the validation dataset $\D_{val}$, and the optimal path $\a^*$ corresponds to the maximum ACC, \ie,
\vspace{-1mm}
\begin{equation}\label{search}
\a^* = \argmax_{\a\in\A}~\mbox{ACC}(\omega_\a^*, \D_{val}),
\vspace{-2mm}
\end{equation}
where $\omega_\a^*\subset \Omega^*$ \wrt path $\a$ in the trained supernet $\N(\Omega^*)$. 

\subsection{Reshaping sampling distribution $p(\A)$}

Current methods assume that the supernet should provide a reasonable ranking over all architectures in $\A$. Thus \textit{all} paths $\a$ are treated equally, and optimized simultaneously \cite{random,face++,fairnas,scarletnas}. 
Then the sampling distribution $p(\A)$ amounts to a uniform distribution $p(\A) = U(\A)$ over $\A$, \ie,
\vspace{-2mm}
\begin{equation}
p(\a) = \frac{1}{|\A|}\mathbb{I}(\a\in \A),
\vspace{-2mm}
\end{equation}
where $\mathbb{I}(\cdot)$ is an indicator function. However, as previously discussed, it is a demanding requirement for the supernet to rank accurately for all paths at the same time. In the huge search space $\A$, there might be some paths of inferior quality. Since the weights are highly shared in the same supernet, training on these weak paths does have negative influence on the evaluation of those potentially-good paths. To alleviate this disturbance, an intuitive idea is to block the training of these weak paths. 

For simplifying the analysis, we assume the search space $\A$ can be partitioned into two subsets $\A_{good}$ and $\A_{weak}$ by an Oracle good but unknown supernet $\N_o$, where
\vspace{-1mm}
\begin{align} \label{partition}
\A = \A_{good} \bigcup \A_{weak}, ~\A_{good} \bigcap \A_{weak} = \emptyset, 
\vspace{-1mm}
\end{align}
and $\A_{good}$ indicates the potentially-good paths while $\A_{weak}$ is for weak paths, \ie,
\begin{align} \label{duibi}
\mbox{ACC}(\a,\N_o,\D_{val}) \geq \mbox{ACC}(\b,\N_o,\D_{val})
\end{align}
holds for all $\a\in \A_{good}, \b\in \A_{weak}$ on validation dataset $\D_{val}$. Then to screen the weak paths and ease the burden of the supernet training, we can just sample from the potentially-good paths $\A_{good}$ instead of all paths $\A$. The sampling distribution $p(\A)$ is equivalently reshaped by truncation on $\A_{good}$, \ie, $\p(\A) = U(\A_{good};\N_o,\D_{val})$ and 
\vspace{-1mm}
\begin{equation} \label{sample_oracle}
p(\a;\N_o,\D_{val}) = \frac{1}{|\A_{good}|}\mathbb{I}(\a\in \A_{good}).
\vspace{-1mm}
\end{equation}
In this way, the supernet is expected to thoroughly get trained on the potentially-good paths and thus give decent performance ranking. Besides, since the valid search space has been shrunken from $\A$ into $\A_{good}$, the training efficiency of supernet is improved accordingly.

\subsection{Greedy path filtering}
Nevertheless, in the supernet training the Oracle supernet $\N_o$ is unknown, thus we can not sample paths according to Eq.\eqref{sample_oracle} since it relies on $\N_o$. In this paper, we propose to use greedy strategy and during training, current supernet $\N_\dag$ is progressively regarded as a proxy of the Oracle $\N_o$.
Thus during the supernet training, we greedily sample paths according to the reshaped sampling distribution given by current $\N_\dag$, namely, $\p(\A) = U(\A_{good};\N_\dag,\D_{val})$. The sampled paths will get optimized, then the supernet is get updated and evolves to a decent performance estimator over $\A_{good}$. 

However, a natural question arises: even given a supernet $\N_\dag$, how can we sample from the shaped distribution $\p(\A) = U(\A_{good};\N_\dag,\D_{val})$? In other words, how can we accurately identify whether a path is from $\A_{good}$ or $\A_{weak}$? Note that the partition of $\A$ is determined by traversing all paths in $\A$ as Eq.\eqref{duibi}, which is not affordable in computation. Since we can not accurately know whether a single path is good or weak, to solve this issue, we propose a multi-path sampling strategy with rejection. 

\begin{figure}[t]
	\centering
	\includegraphics[width=0.6\linewidth]{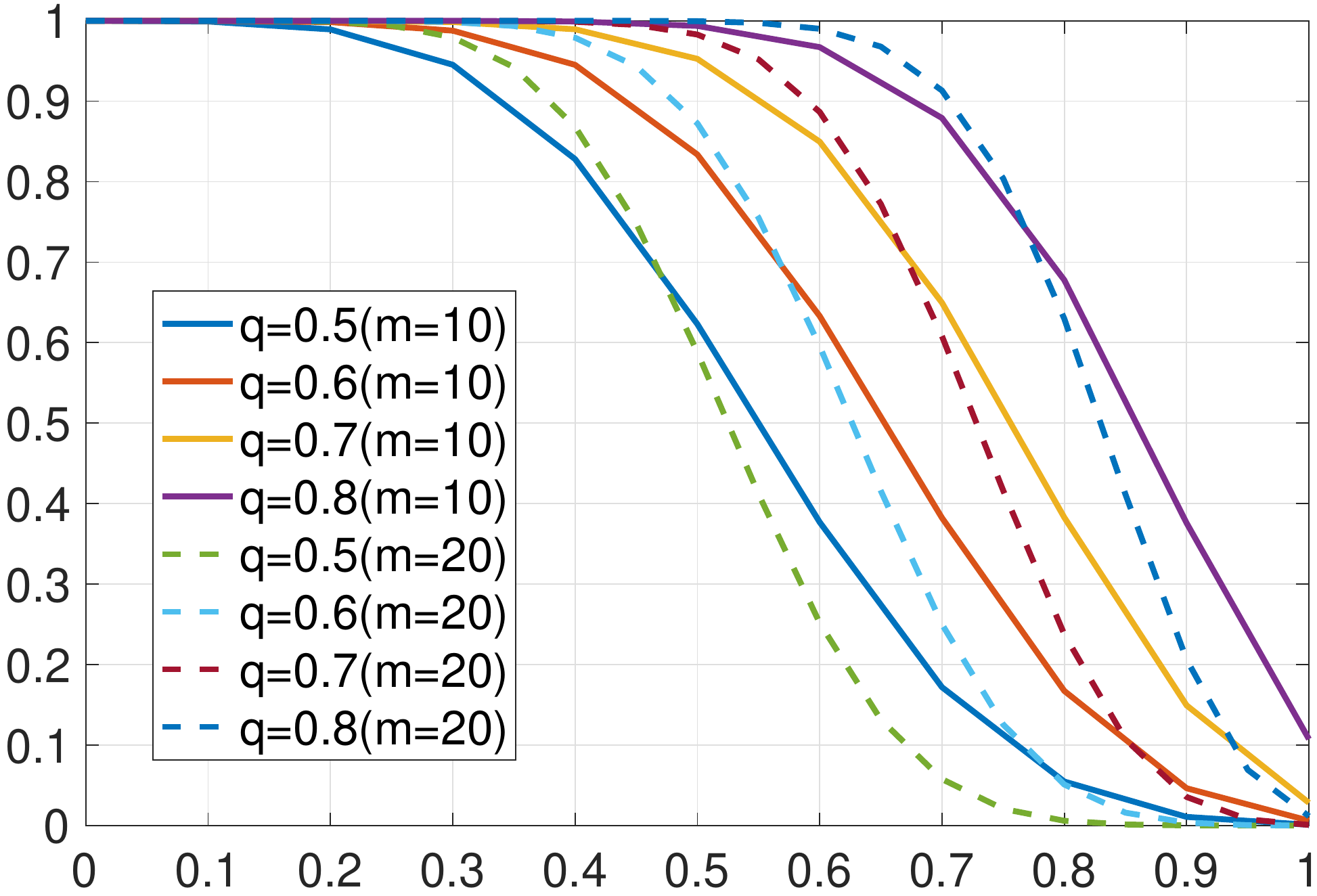}
	\vspace{-3mm}
	\caption{Probability of sampling at least $k$ potentially-good paths out of $m$ paths. X-axis: $r=k/m$. $q=|\A_{good}|/|\A|$.}
	\vspace{-5mm}
	\label{fig:pr}
\end{figure}

Suppose we uniformly sample a path from $\A$, then it amounts to be sampled from $\A_{good}$ with probability $q=|\A_{good}|/|\A|$, and sampled from $\A_{weak}$ with probability $1-q$. In this way, if we sample multiple paths independently at a time, we have the following results based on binomial distribution. 
\begin{theorem}\label{theorem:pr}
	If $m$ paths are sampled uniformly i.i.d. from $\A$, and $\A_{good}$ and $\A_{weak}$ are defined as Eq.\eqref{partition} and Eq.\eqref{duibi} based on supernet $\N_\dag$, then it holds that at least $k~(k\leq m)$ paths are from $\A_{good}$ with probability
	\begin{equation}
	\sum_{j=k}^m \mathbb{C}_m^j q^j (1-q)^{m-j},
	\vspace{-1mm}
	\end{equation}
	where $q=|\A_{good}|/|\A|$. 
\end{theorem}
From Theorem \ref{theorem:pr}, we can see by sampling $m$ paths, the probability that at least $k$ paths are from $\A_{good}$ is very high when the 
proportion of potentially-good paths $q$ is medially large or $k$ is medially small (see Figure \ref{fig:pr}). For example, if we conservatively assume 60\% paths have the potential to be good (\ie, $q=0.6$), we will have 83.38\% confidence to say at least 5 out of 10 paths are sampled from $\A_{good}$. In this way, based on the definition of Eq.\eqref{partition} and Eq.\eqref{duibi}, we just rank the sampled $m$ paths using validation data $\D_{val}$, keep the Top-$k$ paths and reject the remaining paths.

However, ranking $m$ paths involves calculation of ACC over all validation dataset $\D_{val}$ as Eq.\eqref{duibi}, which is also computationally intensive during the supernet training.\footnote{For example, the size of $\D_{val}$ on ImageNet dataset is 50k.} In fact, in our multi-path sampling, what we care about is the obtained ranking; we empirically find that it suffices to rank based on the loss $\ell$ (\eg, cross entropy loss for classification) over a surrogate subset of $\D_{val}$ (\eg, 1k images on ImageNet dataset), denoted as $\hat{\D}_{val}$. The consistency between this rank and that given by ACC on all $\D_{val}$ is fairly significant. More details and analysis refer to the ablation studies in Section \ref{ablation1}. Then the sampling works as Algorithm \ref{alg:sampling}.

\begin{algorithm}[tb]
	\caption{Greedy path filtering w.t/w.o candaidate pool.}
	\label{alg:sampling}
	\begin{algorithmic}[1]
		\REQUIRE{supernet $\N$ with parameter $\Omega$,validation data $\D_{val}$, number of sampled multiple paths $m$, number of kept paths $k$, candidate pool $\P$ with sampling probability $\epsilon$.}
		\IF{without candidate pool $\P$}
		\STATE sample $m$ paths $\{\a_i\}_{i=1}^m$ i.i.d. \wrt  $\a_i \sim U(\A)$
		\ELSE
		\STATE sample $m$ paths $\{\a_i\}_{i=1}^m$ i.i.d. \wrt  $\a_i \sim (1-\epsilon)\cdot U(\A) + \epsilon\cdot U(\P)$
		\ENDIF
		\STATE randomly sample a batch $\hat{\D}_{val}$ in $\D_{val}$ 
		\STATE evaluate the loss $\ell_i$ of each path $\a_i$ on $\hat{\D}_{val}$
		\STATE rank the paths by $\ell_i$, and get Top-$k$ indexes $\{t_i\}_{i=1}^k$
		\STATE return $k$ paths $\{\a_{t_i}\}_{i=1}^k$ and filter the rest
	\end{algorithmic}
\end{algorithm}

As a result, path filtering can be efficiently implemented for it can run in a simple feed-forward mode (\eg, \texttt{eval()} mode in Pytorch) on a small portion of validation data. In this sense, we block the weak paths greedily during the supernet training. And the validation data $\hat{\D}_{val}$ acts as a rough filter to prevent the training of those low-quality or even harmful paths, so that the supernet can get sufficient training on those potentially-good ones.

\section{Proposed Approach: GreedyNAS}
In this section, we formally illustrated our proposed NAS method (a.k.a. GreedyNAS) based on a greedy supernet. Our GreedyNAS is composed with three procedures, \ie, supernet training, searching paths and retraining the searched optimal path. The last retraining corresponds to  conventional training a given network. We mainly elaborate the first two as follows.

\subsection{Greedy training of supernet}
As previously discussed, we propose to maintain a greedy supernet during its training. By doing this, we gradually approximate the sampling $\p(\A) = U(\A_{good};\N_\dag,\D_{val})$ by keeping the Top-$k$ paths and filtering the bottom $m-k$ paths by evaluating using $\hat{\D}_{val}$. Then those weak paths are prevented from getting trained, which allows the supernet to focus more on those potentially-good paths and switch its training space from $\A$ into $\A_{good}$.

\begin{algorithm}[tb]
	\caption{Greedy training of supernet.}
	\label{alg}
	\begin{algorithmic}[1]
		\REQUIRE{supernet $\N$ with parameter $\Omega$, training data $\D_{tr}$, validation data $\D_{val}$, number of sampled multiple paths $m$, number of kept paths $k$, max iteration $T$, training data loader $D$}
		\STATE initialize candidate pool $\P=\emptyset$, 
		\STATE set a Scheduler of pool sampling probability $\epsilon$ 
		\FOR{$\tau =1,..,T/k$}
		\STATE get the pool sampling probability $\epsilon$ by Scheduler
		\STATE sample $k$ paths $\{\a_{t_i}\}_{i=1}^k$ out of $m$ paths using Algorithm \ref{alg:sampling} with pool sampling probability $\epsilon$ 
		\STATE update candidate pool $\P$ using $\{\a_{t_i}\}_{i=1}^k$ 
		\FOR{$i =1,..,k$}
		\STATE get a training batch from $D$
		\STATE update the weights $\omega_{\a_{t_i}}$ of path $\a_{t_i}$ using gradient-based optimizer
		\ENDFOR 
		\ENDFOR
	\end{algorithmic}
\end{algorithm}

\subsubsection{Training with exploration and exploitation}
After the greedy path filtering, we have actually identified some potentially-good paths, which amount to some empirically-good ones given by current supernet. Then to further improve the training efficiency, inspired by the Monte Carlo tree search \cite{kocsis2006bandit} and deep Q-learning (DQN) \cite{mnih2013playing}, we propose to train the supernet with an exploration and exploitation (E-E) strategy by reusing these paths.

Concretely,  we introduce a \textit{candidate pool} $\P$ to store the potentially-good paths discovered during training. Each path $\a$ is represented as a tuple of operation choices. Besides, each $\a_i$ is also allocated with an evaluation loss $\ell_i$. The candidate pool is thus formulated as a fixed-size ordered queue with priority $\ell$. With more potentially-good paths involved, the candidate pool can be maintained by a min-heap structure in real time. 

As a result, we can conservatively implement \textit{local search} by sampling from the candidate pool since it consists of a smaller number (but promising) of paths. However, this greedy \textit{exploitation} brings in the risks of losing path diversity for the training. In this way, we also favor a \textit{global search} with the hope of probing other promising paths that are yet to be sampled and get trained, which can be easily fulfilled by uniform sampling from $\A$. For achieve a balanced trade-off of exploration and exploitation, we adopt a typical $\epsilon$-sampling policy, \ie, implementing uniform sampling both from $\A$ and pool $\P$ (line 4 of Algorithm \ref{alg:sampling}), 
\begin{equation}
\a \sim (1-\epsilon)\cdot U(\A) + \epsilon\cdot U(\P),
\end{equation}
where $\epsilon\in[0,1]$ indicates the probability of sampling from the pool $\P$. Note that candidate pool runs through the training process of supernet; however, it might be not reliable at first since the priority $\ell$ is calculated based on a much less-trained supernet. In this case, we propose to actively anneal the pool sampling probability $\epsilon$ from 0 to a pre-defined level. In our experiment, we find $\epsilon=0.8$ will be a good option. 

Training with exploration and exploitation encourages the supernet to refine the already-found good paths as well as probing new territory for more better paths. Besides, it actually also contributes to our greedy path filtering by improving our filtering confidence. Basically, the collected potentially-good paths can be regarded as a subset of $\A_{good}$, then sampling from $\P$ amounts to increasing the probability $q$ of Theorem \ref{theorem:pr} into 
\begin{equation}
q = \epsilon + (1-\epsilon) |\A_{good}|/|\A|,
\end{equation}
which refers to the proportion of potentially-good paths. For example, assume we evenly sample from $\P$ or $\A$ ($\epsilon=0.5$), then the probability of sampling at least 5 good paths out of 10 paths will rise from $83.38\%$ to $99.36\%$ according to Theorem \ref{theorem:pr}. Comparing reducing $r=k/m$ to increase the sampling confidence, sampling with $\P$ is almost cost-neglectable since we only need to maintain a min-heap. The supernet thus gradually shifts its training from $\A_{good}$ more to $\P$, and the training efficiency will be further improved accordingly.

\begin{figure}[t]
	\centering
	\includegraphics[width=0.95\linewidth]{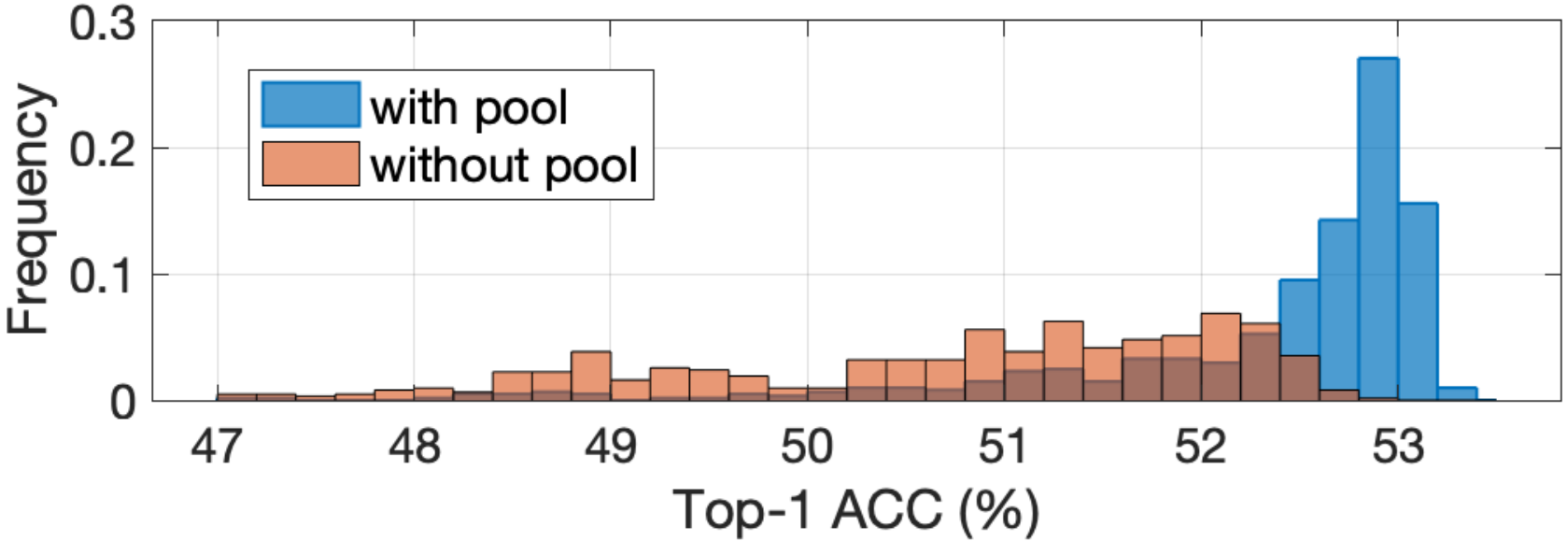}
	\vspace{-3mm}
	\caption{Histogram of accuracy of searched paths on supernet by evolutionary searching method (with or without candidate pool).}
	\vspace{-5mm}
	\label{fig:population}
\end{figure}

\subsubsection{Stopping principle via candidate pool}
Different to conventional networks, a supernet serves as a performance estimator and it is difficult to judge when it is trained well. Current single-path NAS methods control its training by manually specifying an maximum epoch number. In our GreedyNAS, however, we propose an adaptive stopping principle based on the candidate pool. 

Candidate pool $\P$ indicates a bunch of best empirical paths, and it is updated dynamically during the supernet training. In consequence, if a supernet is trained well, the pool $\P$ should tend to be steady. This \textit{steadiness} can be measured by the update frequency $\pi$ of candidate pool $\P$, \ie,
\begin{equation}
\pi := \frac{|\P_t \bigcap \P|}{|\P|} \leq \alpha,
\end{equation}
where $\P_t$ refers to the old $\P$ in previous $t$ iterations. Thus smaller $\pi$ implies that fewer new paths are involved in the pool $\P$ within $t$ iterations, and $\P$ is more steady. Given a certain tolerance level $\alpha$\footnote{$\alpha=0.08$ suffices in our experiment.}, when the update frequency $\pi$ is less than $\alpha$, we believe the supernet has been trained enough, and its training can be stopped accordingly.

\begin{table*}[t]
	\caption{Comparison of classification performance and supernet training efficiency \wrt different searching methods on ImageNet dataset under same search space. \#optimization means the accumulated \#examples calculated for a whole optimization step, while \#evaluation is for that of forward evaluation. corrected \#optimization is based on our statistics that cost of a whole optimization step is 3.33 times larger than that of forward evaluation. Details of calculation refer to supplementary materials.}
	\label{tbl:samespace}
	\centering
	\small
	\vspace{-3mm}
	\begin{tabular}{r||c|c|c|c||c|c|c} 
		\multirow{2}{*}{Methods} & \multicolumn{4}{c||}{performance}  & \multicolumn{3}{c}{supernet training efficiency} 	\\	\cline{2-8}
		~& Top-1 (\%)&FLOPs &  latency & Params&\#optimization & \#evaluation & corrected \#optimization\\ \hline
		Proxyless-R (mobile) \cite{proxylessnas}&74.60&320M& 79 ms &4.0M&-&-&-\\ \hline
		Random Search&74.07&321M&69 ms&3.6M& 1.23M$\times$120&-&147.6M\\
		Uniform Sampling \cite{face++}& 74.50&326M& 72 ms &3.8M&1.23M$\times$120&-&147.6M\\ 
		FairNAS-C \cite{fairnas}& 74.69&321M&75 ms&4.4M&1.23M$\times$150&-&184.5M\\ \hline
		Random Search-E&73.88&320M&91 ms& 3.7M &1.23M$\times$73&-&89.8M\\
		Uniform Sampling \cite{face++}-E&74.17 &320M&94 ms&3.6M&1.23M$\times$73&-&89.8M\\ \hline 
		GreedyNAS (FLOPs$\leq322$M) &\textbf{74.85} &320M& 89 ms &3.8M&1.23M$\times$46&2.40M$\times$46&89.7M\\ 
		GreedyNAS (latency$\leq80$ms) & \textbf{74.93}&324M  & 78 ms &4.1M&1.23M$\times$46&2.40M$\times$46&89.7M \\ \hline
	\end{tabular}	
	\vspace{-5mm}
\end{table*}

\subsection{Searching with candidate pool}
After the supernet is trained, we can use supernet to evaluate the quality (ACC) of each path on validation dataset $\D_{val}$, and search the optimal path $\a^*$ as Eq.\eqref{search}. However, enumerating all paths in $\A$ is prohibitively computation-intensive. One remedy is by dint of evolutionary algorithms \cite{face++} or reinforced version (\eg, MoreMNAS \cite{chu2019multi}), which takes the supernet as an off-the-shelf evaluator. In our paper, we adopt the multi-objective NSGA-\Roman{2} \cite{deb2002fast} algorithm for searching, where the hardware constraint can be easily integrated in the evolution process. If a path violates the pre-defined hardware constraint (\eg, under 330 FLOPs), we just ditch it for good. 

Besides, evolutionary algorithms need to initialize population with size $N_{pop}$ before implementing iterative mutation and crossover. Current methods usually random sample $N_{pop}$ paths under the constraint as initial population. In contrast, our method makes the initialization with the help of candidate pool $\P$, and select its Top-$\N_{pop}$ paths instead. As Figure \ref{fig:population} shows, searching with candidate pool can boost the evolutionary performance for supplying a good initial population. The ACC of searched paths using candidate pool is on average higher than that using random initialization. More details of our searching algorithm refer to the supplementary materials.

\section{Experimental Results} 

\subsection{Configuration and settings}
\textbf{Dataset.} We conduct the architecture search on the challenging ImageNet dataset \cite{imagenet}. As \cite{proxylessnas}, we randomly sample 50,000 images (50 images per class) from training dateset as the validation dataset ($|\D_{val}|=50$K), and the rest of training images are used for training. Moreover, we use the original validation dataset as the test dataset to report the accuracy performance. 

\textbf{Search space.} Following \cite{proxylessnas,fairnas}, we adopt the same macro-structure of supernet for fair comparison as shown in Table 5 (see supplementary materials). Moreover, we use MobileNetV2 inverted bottleneck \cite{mobilenetv2} as the basic building block. For each building block, the convolutional kernel size is within $\{3,5,7\}$ and expansion ratio is selected in $\{3,6\}$. An identity block is also attached for 
flexible depth search. As a result, with 21 building blocks, the search space is of size $(3\times2+1)^{21} = 7^{21}$. In addition, we also implement searching on a larger space by augmenting each building block with an squeeze-and-excitation (SE) option. The size of the larger search space is thus $13^{21}$. 

\textbf{Supernet training.} For training the supernet, Algorithm \ref{alg:sampling} is adopted to sample $10$ paths and filter $5$ paths. We randomly sample $1000$ images  (1 image per class) from the validation dataset for evaluating paths in Algorithm \ref{alg:sampling}. For training each path, we use a stochastic gradient descent (SGD) optimizer with momentum 0.9 and Nesterov acceleration. The learning rate is decayed with cosine annealing strategy from initial value $0.12$. The batch size is $1024$. As for candidate pool, we empirically find 1000 is a good option for pool size $|\P|$, which approximates the amount of paths involved in one epoch. The candidate sampling probability $\epsilon$ is linearly increased from $0$ to $0.8$. Instead of specifying an epoch number \cite{face++,fairnas},  we use the proposed principle to stop the supernet training with tolerance $\alpha=0.08$.

\begin{table*}[t]
	\caption{Comparison of searched architectures \wrt different state-of-the-art NAS methods. \dag: searched on CIFAR-10, $\ddagger$: TPU, $\star$: reported by \cite{face++}.} 
	\label{tbl:sota}
	\centering
	\small
	\vspace{-4mm}
	\begin{tabular}{r||cc||ccc||ccc} 
		Methods& \tabincell{c}{Top-1\\ (\%)}&\tabincell{c}{Top-5\\ (\%)} & \tabincell{c}{FLOPs\\ (M)} &\tabincell{c}{latency\\ (ms)}&\tabincell{c}{Params \\ (M)}&\tabincell{c}{Memory cost}&\tabincell{c}{training cost \\ (GPU days)}&\tabincell{c}{search cost\\ (GPU days)} \\ \hline
		SCARLET-C \cite{fairnas}&  75.6&92.6&280& 67 & 6.0 &single path&10&12\\ 
		MobileNetV2 1.0 \cite{mobilenetv2} &  72.0& 91.0& 300& 38 &3.4 & - & - & - \\
		MnasNet-A1 \cite{mnasnet}& 75.2 & 92.5& 312 & 55 &3.9&single path + RL & $288^\ddagger$ &-\\
		GreedyNAS-C &\textbf{76.2 }&92.5&284& 70 & 4.7&single path&7&$<1$\\ \hline	
		Proxyless-R (mobile) \cite{proxylessnas}&74.6&92.2&320&79&4.0&two paths&$15^\star$&-\\
		FairNAS-C \cite{fairnas}&  74.7&92.1&321& 75 & 4.4 &single path&10&2\\ 
		Uniform Sampling \cite{face++}& 74.7&-&328& - & - &single path&12&$<1$\\
		SCARLET-B \cite{fairnas}&  76.3&93.0&329& 104 & 6.5 &single path&10&12\\ 
		GreedyNAS-B &\textbf{76.8} &93.0&324& 110 &5.2 &single path&7&$<1$\\ \hline
		SCARLET-A \cite{fairnas}&  76.9&93.4&365& 118 & 6.7 &single path&10&12\\  
		EfficientNet-B0 \cite{tan2019efficientnet} & 76.3 & 93.2& 390& 82 & 5.3 &single path& - & - \\ 
		DARTS \cite{darts} & 73.3& 91.3& 574& -&4.7&a whole supernet&$4^\dag$&-\\  
		GreedyNAS-A &\textbf{77.1} &93.3&366& 77&6.5 &single path&7&$<1$\\ \hline
	\end{tabular} 
	\vspace{-5mm}
\end{table*}

\textbf{Evolutionary searching.} For searching with NSGA-\Roman{2} \cite{deb2002fast} algorithm, we set the population size as $50$ and the number of generations as $20$. The population is initialized by the candidate pool $\P$ while other comparison methods use random initialization. During searching, we use constraint of FLOPs or latency. All our experiments use Qualcomm\textregistered\  Snapdragon™ 855 mobile hardware development kit (HDK) to measure the latency. 

\textbf{Retraining.} To train the obtained architecture, we use the same strategy as \cite{proxylessnas} for search space without SE. As for the augmented search space, we adopt a RMSProp optimizer with $0.9$ momentum as Mnasnet \cite{mnasnet}. Learning rate is increased from $0$ to $0.064$ in the first $5$ epochs with batch size $512$, and then decays $0.03$ every $3$ epochs. Besides, exponential moving average is also adopted with decay $0.999$.

\subsection{Performance comparison with state-of-the-art methods}
\textbf{Searching on same search space.} For fair comparison, we first benchmark our GreedyNAS to the same search space as \cite{proxylessnas} to evaluate our superiority to other Single-path One-shot NAS methods. We also cover a baseline method Random Search, which shares the same supernet training strategy with Uniform Sampling \cite{face++}; but during search, instead of using evolutionary algorithms it randomly samples 1000 paths, and retrains the rank-1 path according to Top-1 ACC on the supernet. As Table \ref{tbl:samespace} shows, when searching with similar 320 FLOPs, our GreedyNAS achieves the highest Top-1 ACC. We further align our searched constraint to latency of 80 ms. Table \ref{tbl:samespace} indicates that with similar latency, GreedyNAS is still consistently superior to other comparison methods. For example, GreedyNAS can search an architecture with 74.93\% Top-1 ACC, enjoying a 0.43\% improvement over uniform sampling, which in a way illustrates the superiority of our greedy supernet to a uniform supernet.

Besides advantages on the classification performance of searched models, we also evaluate our superiority in terms of supernet training efficiency. Since the main differences of our GreedyNAS and other Single-path One-shot NAS methods lie in the supernet training, we report in Table \ref{tbl:samespace} the supernet training cost. To eliminate the efficiency gap due to different implementation tools (\eg, GPU types, dataloader wrappers), we calculated the accumulated number of images involved in a whole gradient-based optimization step, \ie, \#optimization in Table \ref{tbl:samespace}. Our GreedyNAS has an additional evaluation process during training, thus we also report the accumulated number of images for forward evaluation, \ie, \#evaluation. For overall efficiency comparison, we empirically find the cost of a whole optimization step is approximately 3.33 times larger than that of a forward evaluation. The corresponding corrected \#optimization is covered accordingly.

From Table \ref{tbl:samespace}, we can see that the training cost of our GreedyNAS is much smaller than that of other comparison methods, which indicates GreedyNAS enjoys significant efficiency in supernet training since it greedily shrinks its training space into those potentially-good paths. Besides, we also implement Random Search and Uniform Sampling using same training cost of GreedyNAS, denoted as \textit{Random Search-E} and \textit{Uniform Sampling-E}, respectively. The results show that with decreased iterations of supernet training, the searched architectures are inferior to those of larger iterations. In contrast, our method can achieve higher accuracy by a large margin (almost 1\%). This implies that GreedyNAS is capable of learning a decent supernet with much less iterations. 

\textbf{Searching on augmented search space.} To comprehensively illustrate our superiority to various state-of-the-art NAS methods, we implement searching by augmenting the current space with an SE option. Moreover, we search the architectures under different FLOPs constraint. But we also report the corresponding latency and parameter capacity to comprehensively analyze the statistics of searched models. As Table \ref{tbl:sota} shows, our GreedyNAS achieves new state-of-the-art performance with respect to different FLOPs and latency levels. For example, with similar FLOPs and latency, GreedyNAS-C has higher Top-1 ACC than the competing SCARLET method by a margin of 0.6\%. Our searched models are visualized in Figure \ref{fig:netarch} (see supplementary materials). It shows smaller network (GreedyNAS-C) tends to select more identity blocks to reduce the FLOPs while larger networks will exploit more SE modules  (GreedyNAS-A\&B) to further improve the accuracy performance. Moreover, GreedyNAS-A adopts more 3$\times$3 kernels to have smaller latency since 3$\times$3 kernels are optimized more maturely in mobile inference framework. We also report our real training cost in Table \ref{tbl:sota} based on Tesla V100 GPU. It shows that GreedyNAS can significantly reduce the training time compared to other NAS methods, which empirically validates the efficiency of GreedyNAS.

\subsection{Ablation studies}

\subsubsection{Effect of evaluation in path filtering} \label{ablation1}
To filter the weak paths, GreedyNAS evaluates each path by a small portion (1000) of validation images as a surrogate for the whole validation dataset (50K images). We first investigate whether this approximation suffices in our experiment. By random sampling 1000 paths from supernet, we examine the correlation of two path orderings, which are generated by ranking the evaluation results using 1000 and 50K validation images, respectively. In Table \ref{tbl:correlation}, we report the widely-used Spearman rho \cite{pirie2004s} and Kendall tau \cite{tau} rank correlation coefficient, which are in the range $[0,1]$ and larger values mean stronger correlation. We also cover three types of supernets, \ie, randomly initialized, trained by uniform sampling and our greedy sampling. 

\begin{figure}[t]
	\centering
	\subfigure
	{\includegraphics[width=0.49\columnwidth]{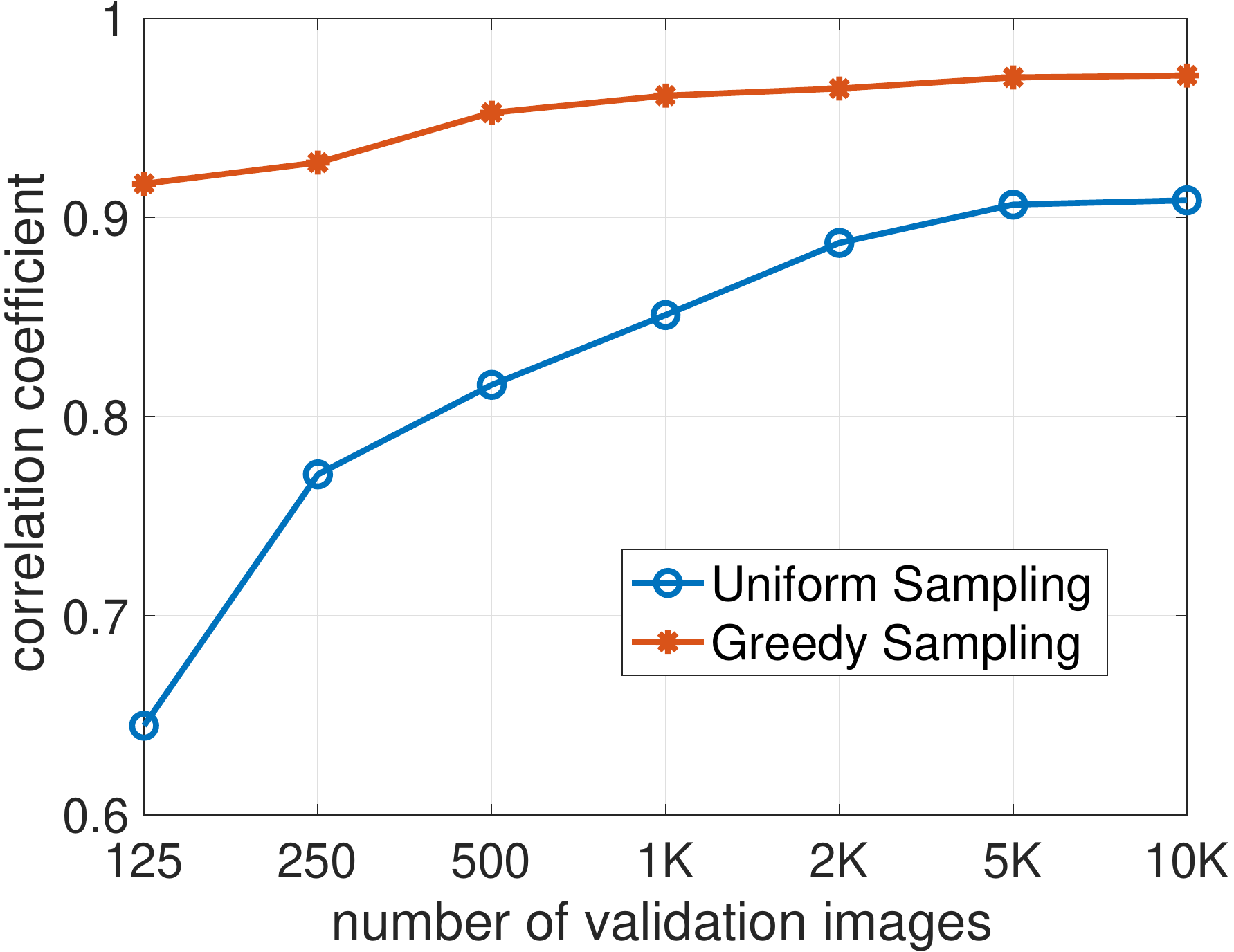}}
	\subfigure
	{\includegraphics[width=0.49\columnwidth]{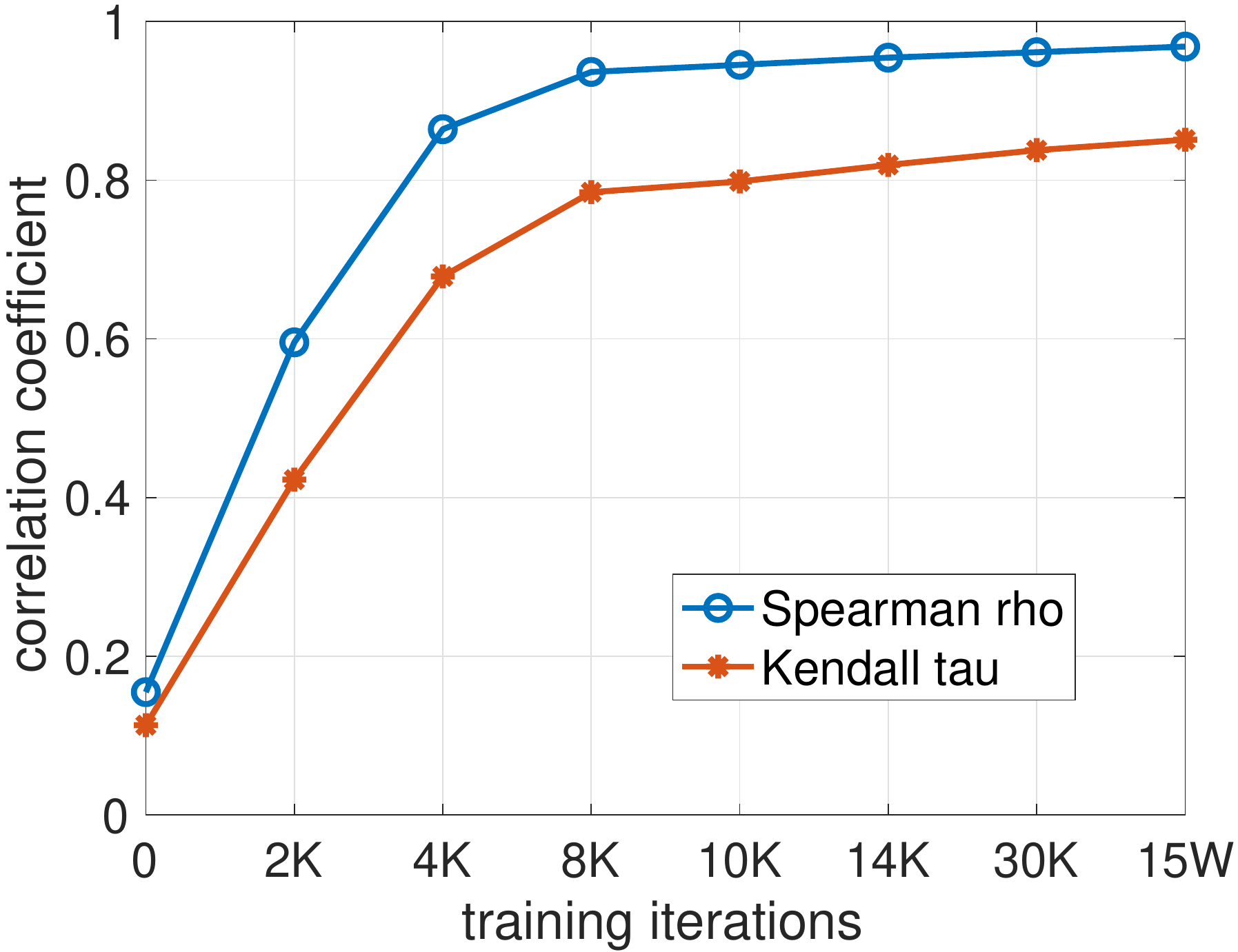}}
	\vspace{-4mm}
	\caption{Rank correlation coefficient of 1000 paths measured by the loss of $N$ validation images and ACC of the whole 50K validation images. Left:  Comparison (Kendall tau) of supernet by uniform and greedy sampling \wrt different number $N$ of evaluation images. Right: $N=1$K \wrt different training iterations of supernet by uniform sampling. }
	\label{fig:ablation}
	\vspace{-4mm}
\end{figure}

From Table \ref{tbl:correlation}, we can see that our greedy supernet achieves fairly high rank correlation coefficient (0.997 and 0.961), which indicates that the ranking of greedy supernet using 1000 validation images is significantly consistent with that of all validation dataset. Moreover, supernet trained with uniform sampling has smaller correlation coefficient, even with different evaluation images (see left Figure \ref{fig:ablation}). This implies in a sense that our greedy supernet is more discriminative since it can use less images to identify whether a path is good or weak. Nevertheless, as left Figure \ref{fig:ablation} shows, too few evaluation images might have weak correlation while too many evaluation images mean greater evaluation cost. But 1000 evaluation images enjoy a balanced trade-off between the rank correlation and evaluation cost. Note that we also report the results \wrt ranking using the ACC of 1000 images, which is smaller than that using loss. This results from that during training the value of loss might be more informative than that of ACC. 

As for the random supernet, the correlation coefficient is fairly small (0.155 and 0.113). This makes sense since the ranking is based on the classification performance; however, a random supernet fails to learn sufficient domain-related information but gives disordered ranking of paths. This smaller correlation coefficient implies that it might be not sensible to implement greedy sampling from a random supernet since the ranking evaluated by 1000 validation images will be rather noisy. In this way, we record the trend of rank correlation coefficients with uniform sampling in right Figure \ref{fig:ablation}. It shows that with more iterations, the correlation coefficients increase and at 10K iteration, they tend to be steady at a high level (\eg, 0.81 for Kendall Tau). As a result, in our GreedyNAS we propose to have a \textit{warm-up} stage by uniform sampling for 10K iterations, so that we can safely use 1000 validation images to evaluate paths.

\begin{table}
	\caption{Rank correlation coefficient of 1000 paths measured by the loss (ACC) of 1K validation images and ACC of 50K validation images \wrt different types of supernets.}
	\label{tbl:correlation}
	\centering
	\small
	\vspace{-3mm}
	\begin{tabular}{p{0.7cm}p{1.6cm}p{0.7cm}||p{0.7cm}p{1.6cm}p{0.7cm}} \hline
		\multicolumn{3}{c||}{Spearman rho} & \multicolumn{3}{c}{Kendall tau} \\ \hline
		random & uniform(ACC) & greedy & random & uniform(ACC) & greedy \\ \hline
		0.155 & 0.968(0.869) & \textbf{0.997} & 0.113 & 0.851(0.699) & \textbf{0.961} \\ \hline
	\end{tabular}	
	\vspace{-5mm}
\end{table}

\subsubsection{Effect of path filtering and candidate pool}
To study the effect of our proposed path filtering and the candidate pool, we implement experiments on the search space without SE. In our GreedyNAS, path filtering is to block the training of weak paths. In contrast, the use of candidate pool is mainly three-fold as shown in Table \ref{tbl:ablation}. First, we can sample from it 
as the exploitation process; second, we can initialize the evolutionary searching with the pool for better paths; third, we can use it to adaptively stop the supernet training. Then we control each factor and obtain 6 variants of GreedyNAS as well as 6 corresponding searched architectures Net1$\sim$Net6. For fair comparison, we search all nets under 330M FLOPs. Besides, if the candidate pool is not used for stopping training, we specify a maximum epoch 60 as \cite{scarletnas}. 

As Table \ref{tbl:ablation} shows, comparing with the baseline Net1 (Net3), Net2 (Net6) achieves 0.28\% (0.41\%) better Top-1 ACC, which indicates that path filtering does contribute to the supernet training, and thus improves the searching results. By involving the candidate pool, Net6 can increase its accuracy from 74.59\% (Net2) to 74.89\%. In specific, initialization with candidate pool in evolutionary algorithms enables to have a 0.18\% gain on Top-1 ACC since it helps to search paths with higher ACC on supernet (also see Figure \ref{fig:population}). Note that stopping by candidate pool usually saves training cost; however, full training with candidate pool (Net5) seems to drop the accuracy a bit (0.05\% \wrt Net6). It might result from that extreme greedy exploitation on the candidate pool harms the supernet training instead. Then the stopping in a sense brings benefits for a more balanced trade-off between exploration and exploitation.


\begin{table}
	\caption{Comparison of accuracy performance of searched paths by GreedyNAS \wrt different usage of path filtering and candidate pool.}
	\label{tbl:ablation}
	\centering
	\footnotesize
	\vspace{-3mm}
	\begin{tabular}{r||c||c|c|c||c} 
		& \multirow{3}{*}{\tabincell{c}{path\\ filtering}} & \multicolumn{3}{c||}{candidate pool}&\multirow{3}{*}{\tabincell{c}{Top-1\\ (\%)}} \\ \cline{3-5} 
		& ~ &sampling&evolutionary&training&~\\ 
		&~&(exploitation)&initialization& stopping & ~\\ \hline
		Net1&-&-&-&-&74.31\\
		Net2& \checkmark&-&-&-&74.59\\
		Net3&-&\checkmark&\checkmark&\checkmark&74.48\\
		Net4&\checkmark&\checkmark&-&\checkmark&74.71\\
		Net5& \checkmark&\checkmark&\checkmark&-&74.84\\
		Net6& \checkmark &\checkmark&\checkmark&\checkmark&74.89\\ \hline
	\end{tabular}	
	\vspace{-5mm}
\end{table}

\section{Conclusion}
Training a supernet is a key issue for Single-path One-shot NAS methods. In stead of treating all paths equally, we propose to greedily focus on training those potentially-good ones. This greedy path filtering can be efficiently implemented by our proposed multi-path sampling strategy with rejection. Besides, we also adopt an exploration and exploitation policy and introduce a candidate pool to further boost the supernet training efficiency. Our proposed method GreedyNAS shows significant superiority in terms of both accuracy performance and training efficiency. 


{\small
	\bibliographystyle{ieee_fullname}
	\bibliography{reference}
}

\onecolumn


\appendix

\setcounter{table}{4}
\setcounter{figure}{4}
\setcounter{algorithm}{2}

\section{Details of evolutionary searching in Section 4.2}
We present the details of our adopted NSGA-\Roman{2} \cite{deb2002fast} evolutionary algorithm in the following Algorithm \ref{alg:evolution}. In our experiment, population size $N_{pop} =50$ and number of generations $T=20$.

\begin{algorithm*}[h]
	\caption{Evolutionary Architecture Search}
	\label{alg:evolution}
	\begin{algorithmic}[1]
		\REQUIRE{supernet $\N$, candidate pool $\P$, population size $N_{pop}$, number of generations $T$, validation data $\D_{val}$, constraints $\C$.}
		\ENSURE{architecture with highest validation accuracy under constraints.}
		\STATE Initialize populations $P_0$ with $\P$ so that $|P_0|=N_{pop}$ and $P_0$ satisfies constraints $\C$. 
		\STATE $E = \emptyset$; \hspace{5cm} \# evaluation set $E$ which stores all evaluated architectures with accuracy
		\FOR{$i = 0, 1,...,T - 1$}
		\STATE $Q_i = make\textendash new \textendash pop(P_i)$; \\
		\hspace{1cm}		\# generate offspring population $Q_i$ using binary tournament selection, recombination, and mutation operators
		\STATE $R_i = P_i \cup Q_i$; 
		\STATE $F_i = fast\textendash non\textendash dominated\textendash sort(R_i)$; \hspace{4.5cm}  \# generate all nondominated fronts of $R_i$
		\STATE $P_{i+1} = \emptyset$ and $j = 0$;  
		\WHILE {$|P_{i+1}| + |F_i|\leq N_{pop}$ }
		\STATE $crowding\textendash distance \textendash assignment(F_i)$; \hspace{5cm} {\# calculate crowding-distance in $F_j$}
		\STATE $P_{i+1} = P_{i+1}\cup F_j$;
		\STATE $j = j + 1$;
		\ENDWHILE
		\STATE $E_i = evaluation\textendash architecture(F_j, D_{val}, \C)$; \hspace{1cm}  \# evaluate architecture with constraints and validation data
		\STATE $E = E \cup E_{i}$ \hspace{11.7cm}  \# extend $E_i$ to $E$
		\STATE $Sort(F_j, E_i)$;  \hspace{9cm} \# sort in descending order using $E_i$
		\STATE $P_{i+1} = P_{i+1} \cup F_j[1:(N_{pop}-|P_{i+1}|)]$; \hspace{3.3cm} \# choose the first $(N_{pop} - |P_{i+1}|)$ elements of $F_j$ 
		\STATE $Q_{i+1} = make\textendash new \textendash pop(P_{i+1})$; \hspace{5.5cm} \# make new population with constraints 
		\ENDFOR
		\RETURN architecture with highest accuracy in $E$
	\end{algorithmic}
\end{algorithm*}


\section{More Experimental Results}

\subsection{Details of (augmented) search space}
The macro-structure of supernet is presented in Table \ref{supernet}, where each operation choice for Choice Block is attached in Table \ref{buildingblock}. 

\begin{table}[h]
	\centering
	\small
	\caption{Macro-structure of supernet. ``input" indicates the size of feature maps for each layer, and ``channels" means for the number of output channels. ``repeat" is for the number of stacking same blocks, and ``stride" is for the stride of first block when stacked for multiple times. ``MB1\_K3" refers to Table \ref{buildingblock}.}
	\label{supernet}
	\begin{tabular}{c|c|c|c|c} \hline
		{input} &{block} & {channels} & {repeat} & {stride} \\ \hline
		$224^2\times 3$ & $3\times 3$ conv & 32 & 1 & 2 \\
		$112^2\times 32$ & MB1\_K3 & 16 & 1 & 1 \\
		$112^2\times 16$ & Choice Block & 32 & 4 & 2 \\
		$56^2\times 32$ & Choice Block & 40 & 4 & 2 \\
		$28^2\times 40$ & Choice Block & 80 & 4 & 2 \\
		$14^2\times 80$ & Choice Block & 96 & 4 & 1 \\
		$14^2\times 96$ & Choice Block & 192 & 4 & 2 \\
		$7^2\times 192$ & Choice Block & 320 & 1 & 1 \\
		$7^2\times 320$ & $1\times 1$ conv & 1280 & 1 & 1 \\ 
		$7^2\times 1280$ & global avgpool & - & 1 & -\\ 
		$1280$ & FC & 1000 & 1& -\\ \hline
	\end{tabular}
\end{table}

\begin{table}[h]
	\centering
	\small
	\caption{Operation choices for each MobileNetV2-based Choice Block in Table \ref{supernet}, where ID means for an identity mapping.}
	\label{buildingblock}
	\begin{tabular}{c|c|c|c} 
		\hline
		{block type} &{expansion ratio} & {kernel} & {SE} \\ \hline
		MB1\_K3 & 1 & 3 & no \\ \hline
		ID &- &-&- \\
		MB3\_K3 & 3 & 3 & no \\
		MB3\_K5 & 3 & 5 & no \\
		MB3\_K7 & 3 & 7 & no \\
		MB6\_K3 & 6 & 3 & no \\
		MB6\_K5 & 6 & 5 & no \\
		MB6\_K7 & 6 & 7 & no \\ \hline 
		MB3\_K3\_SE & 3 & 3 & yes \\
		MB3\_K5\_SE & 3 & 5 & yes \\
		MB3\_K7\_SE & 3 & 7 & yes \\
		MB6\_K3\_SE & 6 & 3 & yes \\
		MB6\_K5\_SE & 6 & 5 & yes \\
		MB6\_K7\_SE & 6 & 7 & yes \\
		\hline
	\end{tabular}
\end{table}

\subsection{Calculating corrected \#optimization in Table 1}
In our GreedyNAS, when equipped with the stopping principle of candidate pool, the supernet training is stopped at approximately 46-th epoch. Thus the accumulated number of examples calculated for a whole optimization step is equal to 
\begin{equation*}
	\mbox{\#optimization=1.23M}\times 46,
\end{equation*}
where 1.23M refers to the quantity of training dataset. As for the path filtering, we evaluate 10 paths based on 1000 validation images, and select 5 paths for training, whose batch size is 1024. In this way, the number of images for evaluation amounts to 
\begin{equation*}
	\begin{split}
		\mbox{\#evaluation} &= \mbox{1.23M}\times \frac{1000}{1024} \times \frac{10}{5} \times 46,\\
		& = \mbox{2.40M}\times 46.
	\end{split}
\end{equation*}
Given our empirical findings that the cost of a whole optimization step is approximately 3.33 times larger than that of a forward evaluation, the corrected \#optimization is thus 
\begin{equation*}
	\begin{split}
		\mbox{corrected \#optimization} &= \mbox{\#optimization} + \mbox{\#evaluation}/3.33,\\
		& = \mbox{1.23M}\times 46 + \mbox{2.40M}\times 46 /3.33,\\
		& = \mbox{89.7M}.
	\end{split}
\end{equation*}

\subsection{Details of rank correlation coefficient}
In ablation study 5.3.1, we use two Spearman rho \cite{pirie2004s} and Kendall tau \cite{tau} rank correlation coefficient to evaluate the correlation of two path orderings, which are generated by ranking the evaluation results using 1000 and 50K validation images, denoted as $\r$ and $\s$,  respectively. Basically, we aim to calculate the correlation of $\r$ and $\s$. 

For Spearman rho rank correlation coefficient, it is simply the Pearson correlation coefficient between random variable $r$ and $s$, if we regard $\r$ and $\s$ as two observation vectors of random variable $r$ and $s$, \ie,
\begin{equation*}
	\rho_{S} = \frac{cov(r,s)}{\sigma_{r}\sigma_s},
\end{equation*}
where $cov(\cdot,\cdot)$ is the covariance of two variables, and $\sigma_{r}~(\sigma_s)$ is the standard deviations of $r~(s)$. Based on observation vectors, it can be more efficiently calculated as 
\begin{equation*}
	\rho_{S} = 1- \frac{6\sum_{i=1}^n (\r_i - \s_i)^2}{n(n^2-1)},
\end{equation*}
where $n=1000$ in our experiment. 

For Kendall tau rank correlation coefficient, it focuses on the pairwise ranking performance. For any pair $(\r_i,\r_j)$ and $(\s_i,\s_j)$, it is said to be \textit{concordant} if $\r_i>\r_j$ and $\s_i>\s_j$ hold simultaneously, or also for $\r_i<\r_j$ and $\s_i<\s_j$. Otherwise, it will be called \textit{disconcordant}. Then the coefficient is calculated as 
\begin{equation*}
	\rho_{K} = \frac{\mbox{\#concordant pairs - \#disconcordant pairs }}{\mbox{\#all pairs}}, 
\end{equation*}
where $\mbox{\#all pairs} = \mathbb{C}_{n}^2$ refers to the total number of pairs. In this way, if two rankings $\r$  and $\s$ are exactly the same, $\rho_{K}$ will be 1 while if the two are completely different (\ie, one ranking is the reverse of the other), $\rho_{K}$ will be -1. According to the definition, it can also be calculated in a closed-form as
\begin{equation*}
	\rho_{K} = \frac{2}{n(n-1)} \sum_{i<j} \sign(\r_i - \r_j) \sign(\s_i - \s_j),
\end{equation*}
where $\sign(\cdot)$ is the sign function.

\subsection{More ablation studies}

\subsubsection{Performance of trained supernet}
To further investigate the performance of the trained supernet, we implement two different searching methods (random search and evolutionary search) on various trained supernet, \ie, greedy supernet, uniform supernet (full training) and uniform supernet-E (same training cost with GreedyNAS). 
The results can be regarded as supplementary for Table 1. 

\begin{table}[h]
	\caption{Comparison of performance on ImageNet dataset of searched architectures \wrt different supernets under same search space.}
	\label{tbl:supernet}
	\centering
	\begin{tabular}{c|c||c|c} 
		supernet & searcher & Top-1 (\%)&FLOPs\\ \hline
		uniform& random&74.07&321M\\
		uniform-E & random&73.88&320M \\
		greedy & random&74.22& 321M \\ \hline
		uniform& evolutionary& 74.50&326M\\ 
		uniform-E& evolutionary&74.17 &320M\\ 
		greedy	&evolutionary&\textbf{74.85} &320M\\ \hline
	\end{tabular}	
\end{table}

\begin{figure}[t]
	\centering
	\includegraphics[width=0.5\linewidth]{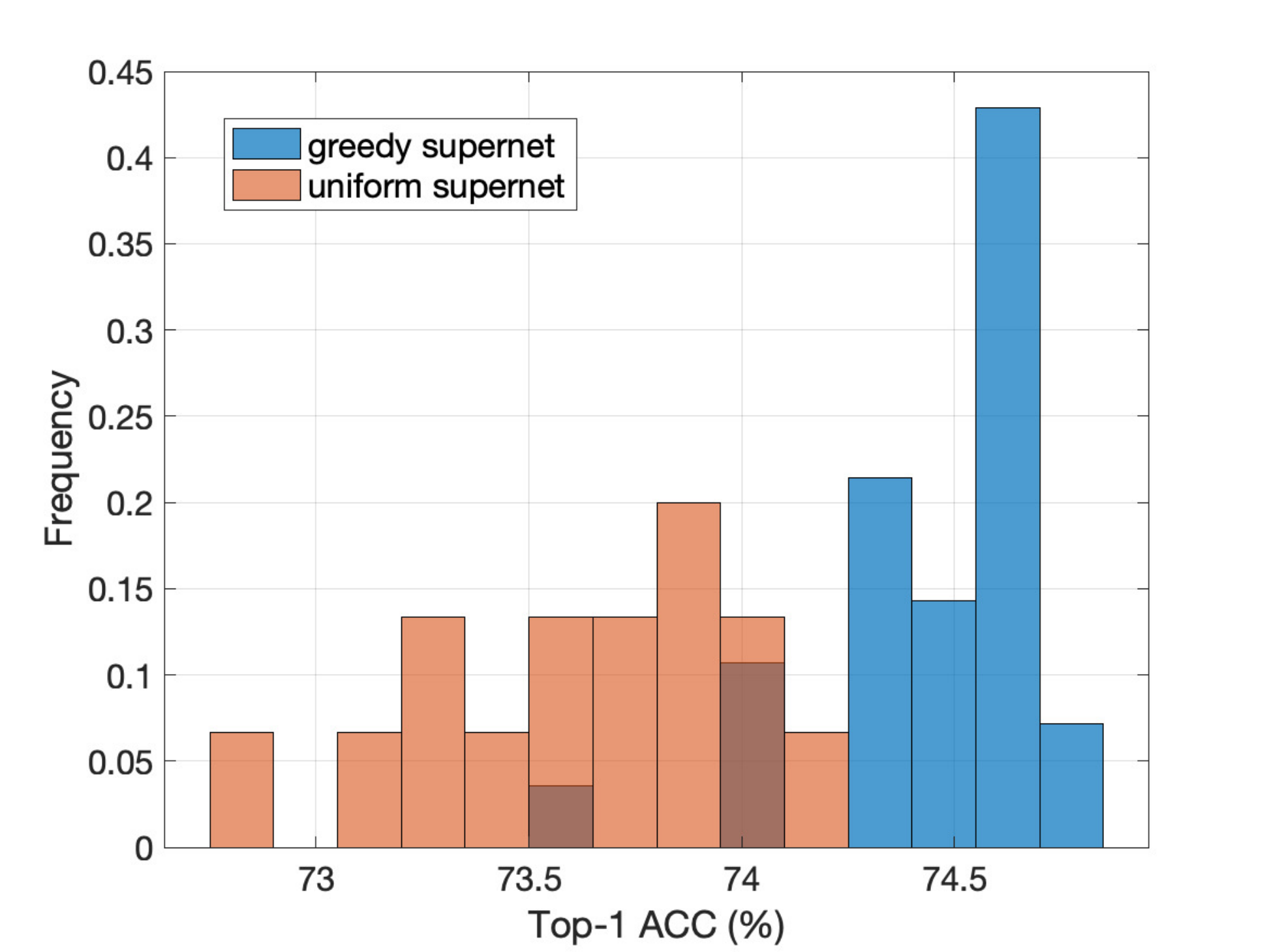}
	\caption{Top-1 accuracy histogram of $30$ systematically sampled paths from $1000$ paths searched by evolutionary algorithm after trained from scratch. }
	\label{fig:supernet}
\end{figure}

From Table \ref{tbl:supernet}, we can see that a greedy supernet improves consistently the classification accuracy in terms of different searchers. This validates the superiority of our greedy supernet since it helps searchers to probe better architectures. Moreover, to comprehensively investigate the effect of supernets, we implement systematic sampling \footnote{\url{https://en.wikipedia.org/wiki/Systematic_sampling}} to sample $30$ paths from $50\times20=1000$ paths, which are discovered by the evolutionary algorithms and ranked according to the accuracy on supernet. Then we retrain these $30$ paths from scratch, and report their distribution histogram in Figure \ref{fig:supernet}. 

As shown in Figure \ref{fig:supernet}, we can see that on average, paths searched with our greedy supernet have higher retraining Top-1 accuracy than that with uniform supernet. This implies that our greedy supernet serves as a better performance estimator, so that those good paths can be eventually identified and searched. 

\newpage
\subsection{Visualization of searched architectures}
\begin{figure*}[t]
	\centering
	\subfigure[GreedyNAS-A]
	{\includegraphics[width=0.2\columnwidth]{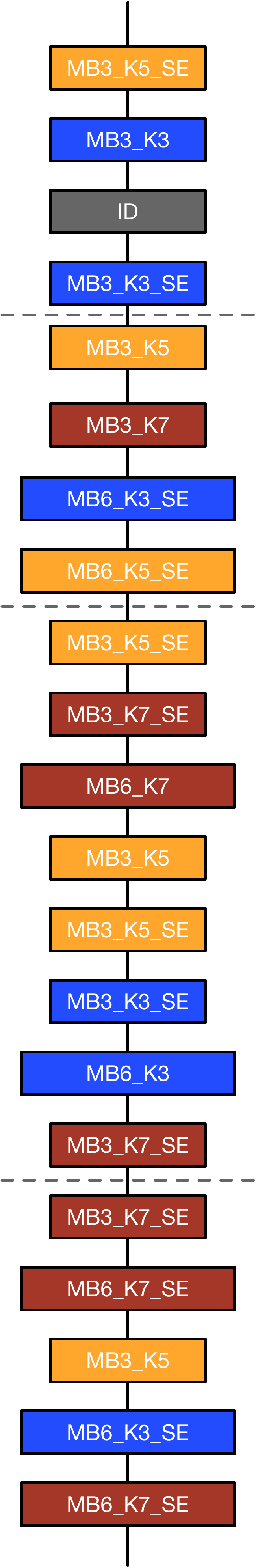}}
	\hspace{5mm}
	\subfigure[GreedyNAS-B]
	{\includegraphics[width=0.2\columnwidth]{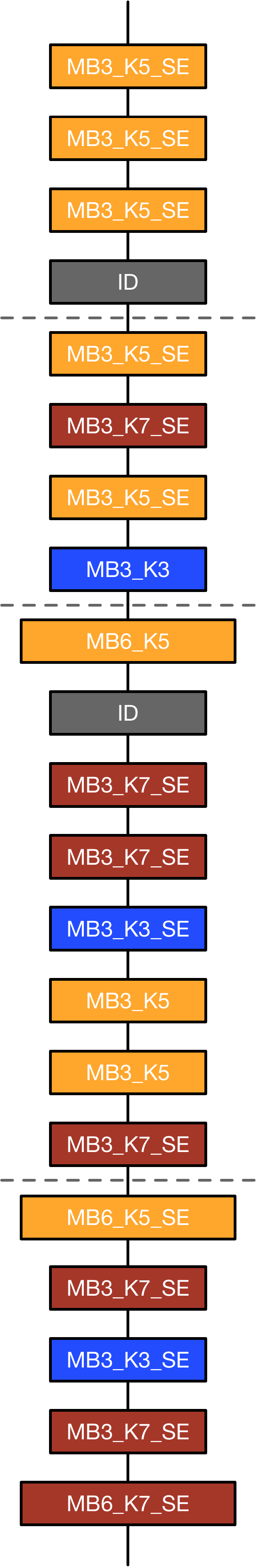}}
	\hspace{5mm}
	\subfigure[GreedyNAS-C]
	{\includegraphics[width=0.2\columnwidth]{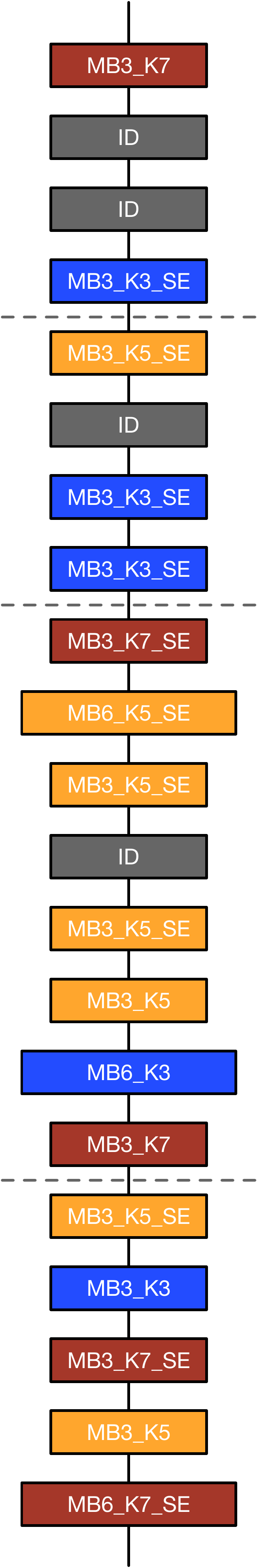}}
	\caption{Visualization of searched architectures by GreedyNAS in Table 2.}
	\label{fig:netarch}
\end{figure*}

We visualize the searched architectures by our GreedyNAS method in Figure \ref{fig:netarch}.  

\end{document}